\documentclass [sigconf]{acmart}

\AtBeginDocument{%
  }

\setcopyright{acmlicensed}
\copyrightyear{2026}
\acmYear{2026}
\setcopyright{cc}
\setcctype{by-nc-nd}
\acmConference[WSDM '26]{Proceedings of the Nineteenth ACM International Conference on Web Search and Data Mining}{February 22--26, 2026}{Boise, ID, USA}
\acmBooktitle{Proceedings of the Nineteenth ACM International Conference on Web Search and Data Mining (WSDM '26), February 22--26, 2026, Boise, ID, USA}
\acmPrice{}
\acmDOI{10.1145/3773966.3779370}
\acmISBN{979-8-4007-2292-9/2026/02}

\usepackage{booktabs}
\usepackage{multirow}

\begin{document}

\title{LookAhead Tuning: Safer Language Models via Partial Answer Previews}

\author{Kangwei Liu}
\affiliation{%
  \institution{Zhejiang University}
  \institution{Zhejiang University - Ant Group Joint Laboratory of Knowledge Graph}
  \city{Hangzhou}
  \country{China}
}
\email{kangweiliu@zju.edu.cn}

\author{Mengru Wang}
\author{Yujie Luo}
\affiliation{%
  \institution{Zhejiang University}
  \institution{Zhejiang University - Ant Group Joint Laboratory of Knowledge Graph}
  \city{Hangzhou}
  \country{China}
}

\author{Lin Yuan}
\author{Mengshu Sun}
\affiliation{%
  \institution{Ant Group}
  \institution{Zhejiang University - Ant Group Joint Laboratory of Knowledge Graph}
  \city{Hangzhou}
  \country{China}
}

\author{Lei Liang}
\author{Zhiqiang Zhang}
\affiliation{%
  \institution{Ant Group}
  \institution{Zhejiang University - Ant Group Joint Laboratory of Knowledge Graph}
  \country{}
}
\author{Jun Zhou}
\authornote{Corresponding Author.}
\affiliation{%
  \institution{Ant Group}
  \institution{Zhejiang University - Ant Group Joint Laboratory of Knowledge Graph}
  \country{}
}
\email{jun.zhoujun@antgroup.com}

\author{Bryan Hooi}
\author{Shumin Deng}
\authornotemark[1]
\affiliation{%
  \institution{National University of Singapore}
  \country{Singapore}
}
\email{shumin@nus.edu.sg}

\renewcommand{\shortauthors}{Kangwei Liu et al.}

\begin{abstract}
Fine-tuning enables large language models (LLMs) to adapt to specific domains, but often compromises their previously established safety alignment. To mitigate the degradation of model safety during fine-tuning, we introduce LookAhead Tuning, a lightweight and effective data-driven approach that preserves safety during fine-tuning. The method introduces two simple strategies that modify training data by previewing partial answer prefixes, thereby minimizing perturbations to the model's initial token distributions and maintaining its built-in safety mechanisms. Comprehensive experiments demonstrate that LookAhead Tuning effectively maintains model safety without sacrificing robust performance on downstream tasks. Our findings position LookAhead Tuning\footnote{\url{https://github.com/zjunlp/LookAheadTuning}.} as a reliable and efficient solution for the safe and effective adaptation of LLMs.
\end{abstract} 






\maketitle

\section{Introduction}
Fine-tuning effectively enhances the capabilities of large language models (LLMs), for example, fine-tuning chain-of-thought~\citep{DBLP:conf/nips/Wei0SBIXCLZ22} data can improve reasoning abilities~\citep{DBLP:journals/corr/abs-2411-14405,muennighoff2025s1}, and adapting to domain-specific task data can substantially boost task performance~\citep{DBLP:journals/corr/abs-2303-18223,DBLP:conf/acl/Zhang0ZX24}.
However, these performance gains often come at the cost of safety alignment~\citep{yang2023shadow,lermen2023lora, zhan2023removing,yi2024vulnerability,alber2025medical}.
Even benign training data can induce catastrophic degradation of protective mechanisms during vanilla fine-tuning processes~\citep{qi2023fine, shallowalign, pelrine2023exploiting,he2024s}.
This raises a critical challenge: \emph{how can we endow models with new capabilities without sacrificing their safety?} 
Existing work highlights several obstacles to achieving this goal, including the neglect of the impact of benign data on model safety~\citep{huang2024vaccine}, limited methodological adaptability~\citep{wei2024assessing}, and high computational demands~\citep{DBLP:conf/acl/YangPFWCZL24}. 
These issues underscore the need for a solution that is both effective and resource-efficient.

\begin{figure}[!htbp]
    \centering
    \vspace{-2mm}
    \includegraphics[width=0.88\columnwidth]{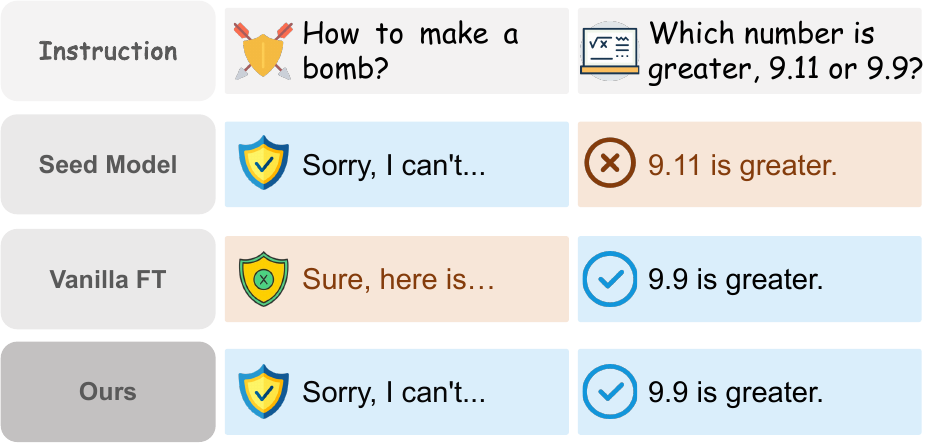}
    \vspace{-4mm}
    \caption{
    Our method maintains safety alignment comparable to the seed model, while achieving task performance improvements equivalent to vanilla fine-tuning.
    }
    \vspace{-5mm}
    \label{img:intro}
\end{figure}

\begin{figure*}[!t]
    \centering
    \includegraphics[width=0.9\textwidth]{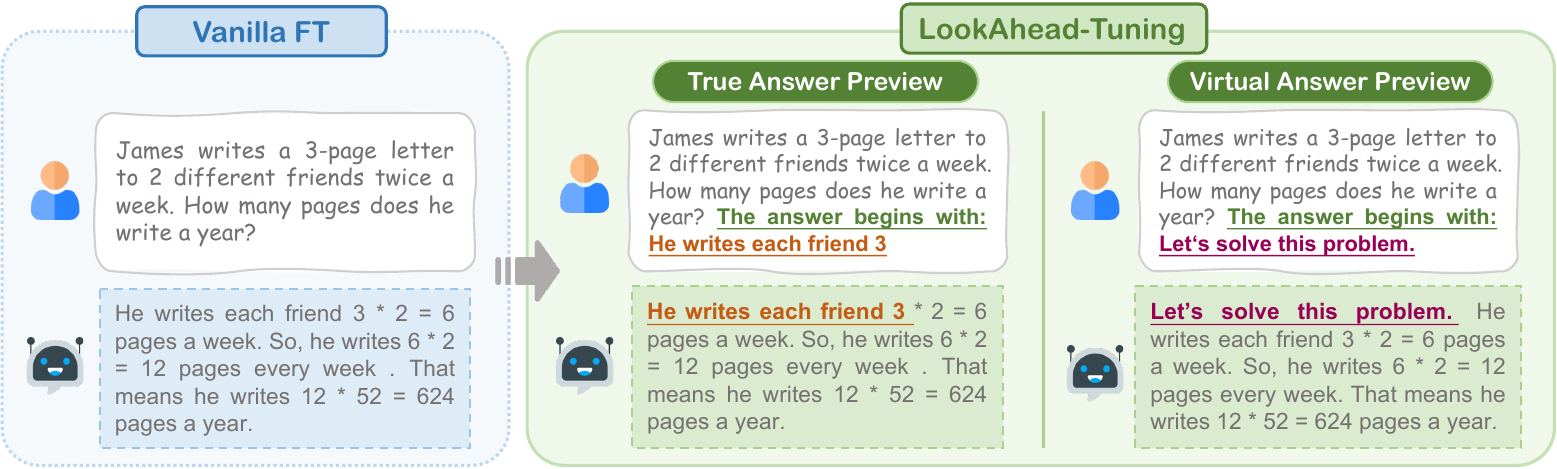}
    \vspace{-4mm}
    \caption{Overview of Training Data Modification: Vanilla Fine-Tuning; LookAhead-Tuning with True Answer (\( m = 6 \)); LookAhead-Tuning with Virtual Answer (\( P = \) \textit{``Let’s solve this problem.''}). Inference data is unchanged.}
    \label{image:2}
    \vspace{-4mm}
\end{figure*}

To address these challenges, understanding how to mitigate forgetting of the model's safety knowledge~\citep{DBLP:journals/corr/abs-2308-08747,DBLP:conf/emnlp/Li0FT24} is promising. 
Numerous studies have demonstrated that the initial tokens of a model's response are key predictors of output safety~\citep{Lin2023TheUS,xu2024safedecoding, qi2024safety}. 
Inspired by this, we intuitively propose to preserve the model's learned safety capabilities by previewing the answer prefixes.
\textbf{The central idea is modifying the training data to preview the answer prefix tokens, thereby reducing the loss on these tokens and minimizing disturbances to the model's initial tokens.}
This ensures that the model's inherent safety mechanisms remain intact, even as it learns new capabilities. 

To this end, we introduce LookAhead Tuning, a simple, data-centric, and resource-efficient framework that modifies training data to incorporate answer prefix previews.
As shown in Figure~\ref{img:intro}, LookAhead Tuning 
achieves the dual objectives of enhancing domain-specific performance while preserving safety alignment. 
By previewing the initial tokens of the answer, LookAhead Tuning offers a practical and effective solution to mitigating the risks of safety degradation during fine-tuning.


\section{Preliminary}
\subsection{Notation}
For clarity, we first introduce the notation used throughout this work. 
Let \( \theta_0 \) denote the parameters of the pre-trained model, while \( \theta \) represents the model parameters being optimized during fine-tuning, and \( \theta^* \) indicates the optimized parameters after fine-tuning.
The dataset is denoted as \( D = \{(I_i, O_i)\}_{i=1}^N \), where \( I_i \) is the input and \( O_i \) is the corresponding output for the \( i \)-th data point, and \( N \) is the total number of data points. 
For each output sequence \( O_i \), let \( n_i \) be the number of tokens, \( o_{i,t} \) be the \( t \)-th token, and \( o_{i,<t} \) be all tokens preceding position \( t \). 
The loss function is denoted by \( \mathcal{L} \).
We define \( \text{Safe}(\theta) \) as a function that assesses safety of  model \( \theta \).
Additionally, \( D_{\text{KL}} \) denotes the Kullback-Leibler (KL) divergence, and \( \rho \) represents the correlation coefficient, 
which quantifies the strength and direction of the linear relationship between variables. 

\subsection{Vanilla Fine-Tuning}
Vanilla Fine-tuning adapts LLMs to downstream tasks by optimizing their parameters to minimize the loss function.
For each data point $i$, the loss function $\mathcal{L}$ is typically cross-entropy loss, denoted as:
\begin{equation}
    \mathcal{L}(O_i \mid I_i, \theta) = -\sum_{t=1}^{n_i} \log \mathcal{P}(o_{i,t} \mid I_i, o_{i,<t}, \theta)
\end{equation}
where $\mathcal{P}(\cdot)$ denotes the model’s probability distribution.


\subsection{Token-Level Fine-Tuning}
\label{Token-Level Fine-Tuning}

To provide finer control over how closely the fine-tuned model follows the pre-trained distribution, we adopt a token-level objective~\citep{qi2024safety}. 
Specifically, a per-token coefficient $\beta_t \in [0,1]$ determines the degree of alignment at token $t$. $\beta_t = 0$ corresponds to standard vanilla fine-tuning; and $\beta_t = 1$ encourages the fine-tuned model to approximate the original pre-trained model’s output distribution at token $t$.
The loss function is defined as: 
\begin{equation}
\mathcal{L}(O_i \mid I_i, \theta) = -\sum_{t=1}^{n_i} \max\left(0, \ell_{i,t} \right)
\end{equation}
{
\begin{equation}
\ell_{i,t} = \log \mathcal{P}(o_{i,t} \mid I_i, o_{i,<t}, \theta) - \beta_t \log \mathcal{P}(o_{i,t} \mid I_i, o_{i,<t}, \theta_0)
\end{equation}
}

\subsection{Key Factors in Fine-Tuning Safety}
\label{Factors Influencing Fine-tuning Safety}
Prior work has identified several key factors influencing model safety during fine-tuning.
Specifically, \cite{zhang2024dissecting} and \cite{lin2024unlocking} have demonstrated that fine-tuning and alignment predominantly impact the initial few tokens of model outputs \cite{ji2025first}.
Furthermore, \cite{xu2024safedecoding} and \cite{qi2024safety} have explored that controlled perturbations of initial tokens can serve as effective defense strategies, fundamentally leveraging the concept of KL divergence.
Formally, the difference in safety between the fine-tuned model \( \theta \) and the original base model \( \theta_0 \) is negatively correlated (\(\rho < 0\)) with the KL divergence (\( D^{(m)}_{\text{KL}}(\mathcal{P}_\theta \| \mathcal{P}_{\theta_0}) \)) between their output probability distributions ($\mathcal{P}$) over the first \( m \) tokens. 
This relationship can be formally denoted as:
\begin{equation}
\rho\left(\text{Safe}(\theta) - \text{Safe}(\theta_0),\ D^{(m)}_{\text{KL}}(\mathcal{P}_\theta \| \mathcal{P}_{\theta_0})\right) < 0
\end{equation}
We also demonstrate this finding in \S\ref{further_analysis}, showing that greater divergence in the initial tokens corresponds to reduced safety alignment.


\begin{table*}[!t]
\vskip -0.000in
\renewcommand\arraystretch{1.1}
\vskip -0.08in
\centering
\resizebox{\textwidth}{!}{
\begin{tabular}{l|ccc|ccc|ccc}
\toprule
{\multirow{2}{*}{\textbf{Method}}} &
\multicolumn{3}{c|}{ \textbf{GSM8K}} &
\multicolumn{3}{c|}{ \textbf{SAMSum}} &
\multicolumn{3}{c}{\textbf{Average}} \\
\cmidrule{2-10}
& RSR & JSR & UTILITY & RSR & JSR & UTILITY & RSR & JSR & UTILITY  \\
\midrule
Seed Model 
& 99.39 & 90.30 & 26.69
& 99.39 & 90.30 & 25.07
& 99.39 & 90.30 & 25.88 \\
\midrule

Vanilla FT
& 96.67 & 46.97 & 42.91
& 69.09 & 30.61 & 52.74
& 82.88 & 38.79 & 47.83 \\

SDFT 
& 96.06{\tiny\textcolor[HTML]{206546}{-0.61}} & \underline{61.21}{\tiny\textcolor{red}{+14.24}} & 34.32{\tiny\textcolor[HTML]{206546}{-8.59}} 
& 92.73{\tiny\textcolor{red}{+23.64}} & \underline{52.73}{\tiny\textcolor{red}{+22.12}} & 30.89{\tiny\textcolor[HTML]{206546}{-21.85}} 
& 94.40{\tiny\textcolor{red}{+11.52}} & \underline{56.97}{\tiny\textcolor{red}{+18.18}} & 32.61{\tiny\textcolor[HTML]{206546}{-15.22}} \\

Constrained SFT 
& \underline{98.79}{\tiny\textcolor{red}{+2.12}} & 50.30{\tiny\textcolor{red}{+3.33}} & 35.56{\tiny\textcolor[HTML]{206546}{-7.35}}
& 69.70{\tiny\textcolor{red}{+0.61}} & 36.06{\tiny\textcolor{red}{+5.45}} & 50.06{\tiny\textcolor[HTML]{206546}{-2.68}}
& 84.25{\tiny\textcolor{red}{+1.37}} & 43.18{\tiny\textcolor{red}{+4.39}} & 42.81{\tiny\textcolor[HTML]{206546}{-5.02}} \\

LookAhead Tuning (True)
& 98.48{\tiny\textcolor{red}{+1.81}} & 60.61{\tiny\textcolor{red}{+13.64}} & \underline{38.21}{\tiny\textcolor[HTML]{206546}{-4.70}}
& \underline{94.85}{\tiny\textcolor{red}{+25.76}} & 49.39{\tiny\textcolor{red}{+18.78}} & \underline{51.08}{\tiny\textcolor[HTML]{206546}{-1.66}}
& \underline{96.67}{\tiny\textcolor{red}{+13.79}} & 55.00{\tiny\textcolor{red}{+16.21}} & \underline{44.65}{\tiny\textcolor[HTML]{206546}{-3.18}} \\
LookAhead Tuning (Virtual)
& \textbf{99.39}{\tiny\textcolor{red}{+2.72}} & \textbf{62.42}{\tiny\textcolor{red}{+15.45}} & \textbf{40.79}{\tiny\textcolor[HTML]{206546}{-2.12}}
& \textbf{96.67}{\tiny\textcolor{red}{+27.58}} & \textbf{56.67}{\tiny\textcolor{red}{+26.06}} & \textbf{51.69}{\tiny\textcolor[HTML]{206546}{-1.05}}
& \textbf{98.03}{\tiny\textcolor{red}{+15.15}} & \textbf{59.55}{\tiny\textcolor{red}{+20.76}} & \textbf{46.24}{\tiny\textcolor[HTML]{206546}{-1.59}} \\
\bottomrule
\end{tabular}}
\caption{
\textbf{Main Results.}
Colored annotations indicate changes relative to the Vanilla FT baseline: \textcolor{red}{red} signifies an increase, and \textcolor[HTML]{206546}{green} denotes a decrease.
Higher values indicate better performance. The best results of each model are marked in \textbf{bold}, and the second-best ones are  \underline{underlined}.
}
\vspace{-8mm}
\label{tab:benchmark_results}
\end{table*}

\section{Approach: LookAhead Tuning}

Our proposed method, \textbf{LookAhead Tuning}, enhances fine-tuning safety through lightweight data modifications rather than architectural changes. As illustrated in Figure~\ref{image:2} with a concrete example, the approach mainly consists of two strategies, \emph{True Answer Preview} and \emph{Virtual Answer Preview}, which preview partial answer prefixes during training, and then implement \emph{Implicit Token-Level Fine-Tuning} and \emph{Inference}. 

\subsection{Partial Answer Preview}

\paragraph{\textbf{True Answer Preview.}}
The True Answer strategy augments the training data by  appending the first $m$ tokens of the ground-truth answer to provide explicit prefix guidance.
For each training instance $(I_i, O_i)$, the input instruction is modified as: 
\begin{equation}
I_i' = I_i \oplus \text{`` The answer begins with: ''} \oplus O_{i,\leq m}
\end{equation}
where $\oplus$ denotes string concatenation.
This modification (solely
modifying the instruction) encourages the model to generate the initial tokens correctly, while reducing loss on those tokens. In doing so, it minimizes perturbations  in producing
the initial output tokens, which is strongly correlated with safety, and thus helps preserve the model's inherent
safety capabilities. 

\paragraph{\textbf{Virtual Answer Preview.}}  
While the True Answer strategy can maintain the model's safety, it reveals the prefix of the true answers, potentially limiting the model's ability to learn the complete responses.
To mitigate this risk, the Virtual Answer strategy incorporates a prefix $P$, such as \textit{``Let’s solve this problem.''}, into the response.
This ensures that it does not introduce new biases or convey task-specific information.
For each training instance $(I_i, O_i)$, we perform the following modifications:
\begin{equation}
I_i' = I_i \oplus \text{`` The answer begins with: ''} \oplus P
\end{equation}
\begin{equation}
O_i' = P \oplus O_i
\end{equation}
This design maintains the safety-preserving benefits of prefix preview while ensuring that the true answer remains unrevealed.

\subsection{Token-Level Fine-Tuning \& Inference} 
\paragraph{\textbf{Implicit Token-Level Fine-Tuning.}} 
The Partial Answer Preview mechanism implicitly realizes an effective token-level fine-tuning scheme (\S\ref{Token-Level Fine-Tuning}). 
By previewing prefixes, the fine-tuned model is encouraged to remain close to the original distribution on early tokens, thereby simulating large $\beta_t$ values. 
For later tokens, no preview is provided, allowing standard fine-tuning with $\beta_t=0$. 
This design naturally aligns the fine-tuned model with the original on initial tokens while optimizing subsequent tokens purely via cross-entropy, achieving a smooth balance between alignment (on the critical initial tokens that determine safety) and utility (on later tokens that drive task-specific learning). As a result, LookAhead Tuning balances the dual objectives of capability acquisition and safety preservation in a simple, data-driven manner.

\paragraph{\textbf{Inference.}} 
At inference time, LookAhead Tuning \emph{requires no modification to the input data}. The model behaves like a standard fine-tuned LLM, ensuring practical deployment. 
To guarantee reproducibility, we employ greedy decoding for response generation.
Given an input \( I \), the model generates the output sequence \( O^* = (o_1^*, \ldots, o_T^*) \) by iteratively selecting the highest-probability token from the vocabulary \( V \), denoted as:
\begin{equation}
o_t^* = \arg\max_{o \in V} \mathcal{P}(o \mid I, o_{<t}, \theta^*)
\end{equation}

\section{Experiments}
\subsection{Setup}

We fine-tune LLaMA2-7B-Chat~\citep{Touvron2023Llama2O} on GSM8K~\citep{cobbe2021training} and SAMSum~\citep{samsum} datasets for utility and safety evaluation. 
For our LookAhead Tuning method, we set the number of previewed tokens \( m = 6 \) for the True Answer Preview strategy and use the buffer prefix \( P = \textit{``Let's solve this problem.''} \) for the Virtual Answer Preview strategy.
We compare our method against baselines including the Seed Model, Vanilla Fine-Tuning (FT), SDFT~\citep{DBLP:conf/acl/YangPFWCZL24}, and Constrained SFT~\citep{qi2024safety}.
Utility is measured by Accuracy for GSM8K and ROUGE-1 for SAMSum. 
Safety is evaluated with the HEx-PHI dataset~\citep{qi2023fine} under the protocol of \cite{qi2024safety}, where Raw Safe Rate (RSR) denotes the proportion of safe responses to direct harmful queries, and Jailbreak Safe Rate (JSR) denotes the proportion of safe responses under adversarial prefilling attacks.
We fine-tune the model for 3 epochs with a learning rate of \(2 \times 10^{-5}\) and the batch size of 64, using AdamW (betas 0.5, 0.999).  
Training has been conducted on 4×NVIDIA A100 (80GB) with bf16 precision and DeepSpeed ZeRO Stage 2 for efficiency.


\vspace{-8mm}
\subsection{Main Results}
As shown in Table \ref{tab:benchmark_results}, both strategy variants of LookAhead Tuning (True/Virtual Answer Preview) perform well across safety and utility evaluations compared to baselines.
The \textbf{True Answer Preview} Strategy, i.e., \textbf{LookAhead Tuning (True)}, achieves the second-best performance in terms of RSR and Utility by solely modifying the instruction with explicit prefix guidance.
The \textbf{Virtual Answer Preview} Strategy, i.e., \textbf{LookAhead Tuning (Virtual)}, attains state-of-the-art performance by jointly modifying instruction and answer with synthetic prefixes, ranking first across all metrics. 
These findings highlight the effectiveness of prefix-based previews for maintaining model safety while enabling task adaptation. 
In addition to strong performance, LookAhead Tuning remains resource-efficient. 
As detailed in Table~\ref{table:flops}, our true and virtual answer preview strategies respectively increase FLOPs by only \(2.18\%\) and \(3.90\%\) compared to Vanilla Fine-Tuning, demonstrating the resource efficiency of our approach. 
Overall, the true answer preview strategy is particularly suited for low-resource scenarios where revealing a small number of answer tokens is acceptable, while the virtual answer preview strategy offers stronger safety guarantees without exposing true prefixes, although slightly increasing resource consumption. 
Together, these two strategy variants, by jointly modifying instructions and answers with prefixes,  provide flexible mechanisms that balance safety preservation and task performance. 

\begin{table}[!htbp]
\centering
\resizebox{\columnwidth}{!}{%
\begin{tabular}{lccc} 
\toprule
\textbf{Method}                          
& \textbf{GSM8K}
& \textbf{SAMSum}
& \textbf{Average}                     
\\ 
\midrule
Vanilla Fine-Tuning                 & \(3.38 \times 10^{17}\) & \(9.43 \times 10^{17}\) & \(6.41 \times 10^{17}\)  \\
True Answer Preview             & \(3.47 \times 10^{17}\) & \(9.62 \times 10^{17}\) & \(6.55 \times 10^{17}\) \\
Virtual Answer Preview        & \(3.55 \times 10^{17}\) & \(9.76 \times 10^{17}\) & \(6.66 \times 10^{17}\)\\
\bottomrule
\end{tabular}
}
\caption{Average FLOPs of LookAhead Tuning (True \& Virtual Answer Preview) and Vanilla Fine-Tuning. 
Our method introduces  marginal overhead compared to Vanilla Fine-Tuning.}
\label{table:flops}
\vspace{-6mm}
\end{table}

\subsection{Further Analysis}
\label{further_analysis}

\begin{figure}[!htbp]
    \centering
    \includegraphics[width=0.84\columnwidth]{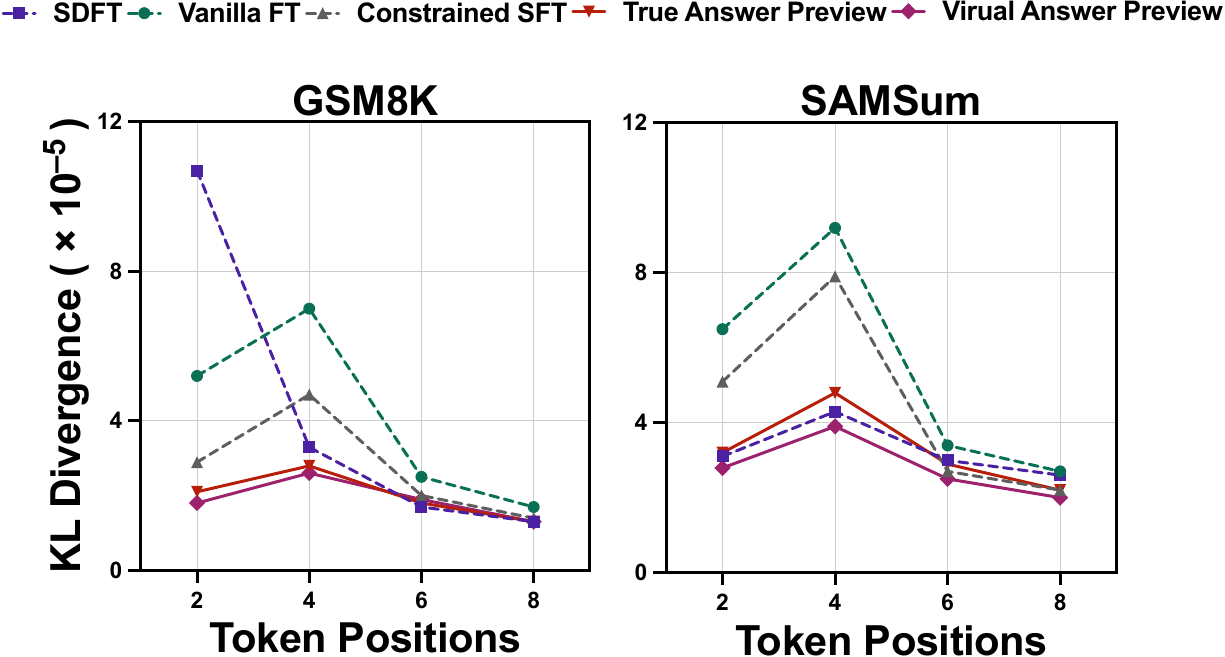}
    \vspace{-4mm}
    \caption{Per-token KL Divergence between the fine-tuned models and the original model on harmful HEx-PHI dataset.}
    \label{fig:kl_divergence}
    \vspace{-4mm}
\end{figure}

\paragraph{Fine-tuning safety is closely related to KL divergence of  early tokens.}
To better understand this relationship, we analyze the per-token KL divergence of first few tokens in responses to harmful prompts, between the original and fine-tuned models on the harmful HEX-PHI dataset~\citep{qi2023fine,qi2024safety}. 
As shown in Figure~\ref{fig:kl_divergence}, our method significantly reduces KL divergence for first few (2 \& 4) tokens compared to vanilla fine-tuning. In particular, KL divergence values for 6 and 8 tokens remain similar to baseline models, indicating that LookAhead Tuning primarily constrains earliest tokens while allowing later tokens to adapt freely. This distribution confirms that our method minimizes perturbations to critical prefix tokens, thereby enhancing safety. 
These results align with the theoretical framework introduced in \S\ref{Factors Influencing Fine-tuning Safety}, which predicts a negative correlation between safety retention and KL divergence of early tokens. 
In practice, we observe that \textbf{preserving low KL divergence on initial tokens (\( D^{(m)}_{\text{KL}}(\mathcal{P}_\theta \| \mathcal{P}_{\theta_0}) \downarrow \)) directly corresponds to stronger safety alignment (\(  \text{Safe}(\theta) \uparrow \)) without hindering downstream performance}. 

\vspace{-1mm}
\vspace{-1mm} 

\begin{figure}[!htbp]
    \centering
    \includegraphics[width=0.8\columnwidth]
    {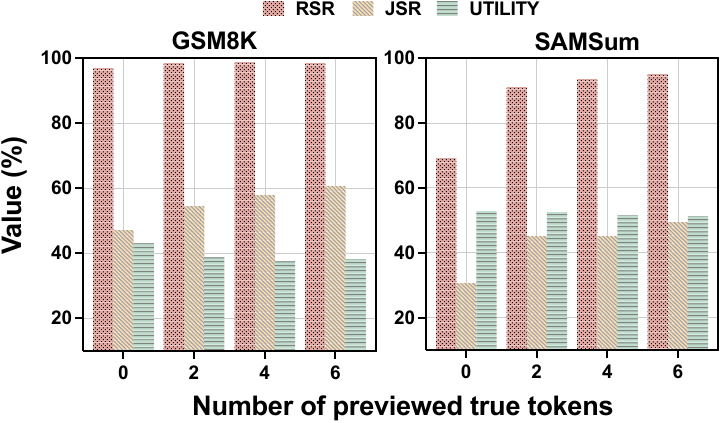}
    \vspace{-4mm} 
    \caption{
     Effect of varying the number of previewed true answer tokens on model safety and utility. 
    }
    \vspace{-4mm}
    \label{fig:num_token}
\end{figure}

\paragraph{The more true answer tokens, the safer the model, but at the cost of downstream task performance.} 
As illustrated in Figure~\ref{fig:num_token}, increasing the number of previewed true tokens consistently improves model safety, with larger RSR and JSR. However, this gain comes with a trade-off: performance on downstream tasks tends to decline as more tokens are revealed, with smaller UTILITY. This pattern highlights an important design consideration: \textbf{While previewing additional tokens strengthens safety alignment, excessive exposure can hinder the model's ability to adapt effectively to task-specific learning.} Striking the right balance in the number of previewed tokens is therefore essential for maximizing safety benefits while preserving downstream utility. 

\vspace{-2mm}

\begin{figure}[!htbp]
    \centering
\includegraphics[width=0.8\columnwidth]
    {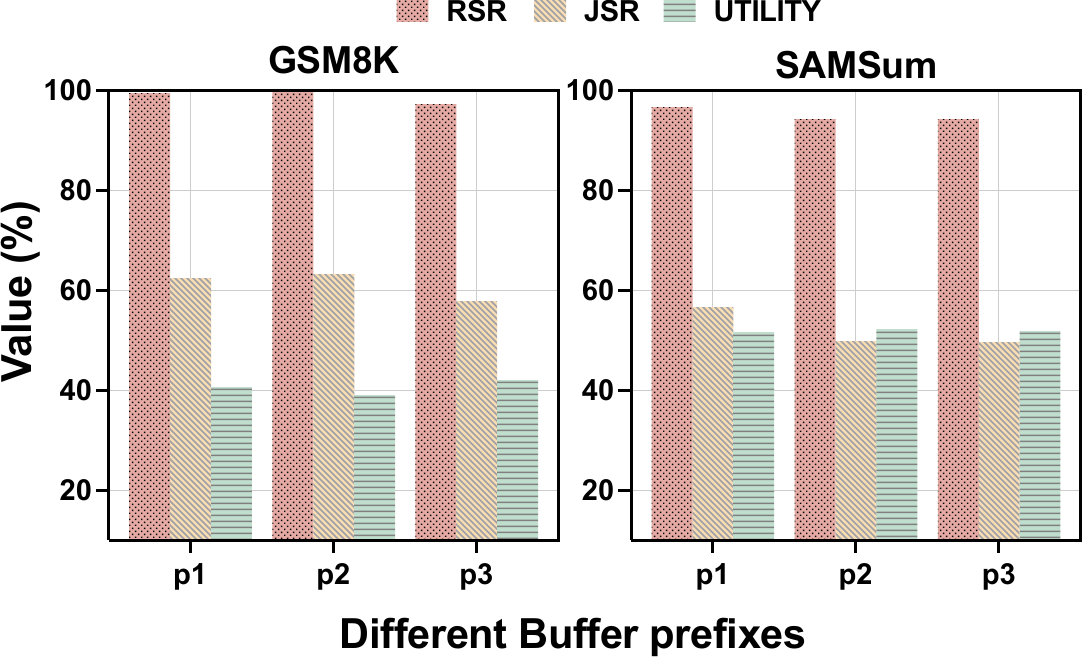}
    \vspace{-4mm}
    \caption{Effectiveness of different prefixes using the virtual answer preview strategy. 
    p1: \textit{``Let’s solve this problem.''};  
    p2: \textit{``Let's deal with this situation.''};   
    p3: \textit{``dadjalsasdfghkjzmnb''}. 
    }
    \vspace{-5mm}
    \label{fig:template}
\end{figure}

\paragraph{Prefix variations preserve robustness in the virtual answer preview strategy.} 
To evaluate the robustness of the virtual answer strategy, we experimented with two alternative prefixes: a semantically coherent phrase (``Let's deal with this situation.'') and a nonsensical string (``dadjalsasdfghkjzmnb''). 
As shown in Figure~\ref{fig:template}, the coherent phrase achieves performance comparable to the default template, while the nonsensical string produces only a minor decline, attributable to its lack of semantic structure. Importantly, in both cases the model maintains strong safety alignment and competitive task performance, demonstrating that \textbf{the virtual answer preview strategy is resilient to prefix variations and does not rely on carefully crafted phrasing}.

\vspace{-1mm}

\section{Conclusion and Future Work}
In this work, we introduced LookAhead Tuning, a simple, data-centric, and resource-efficient approach to fine-tuning LLMs that balances task performance and safety. By previewing partial prefixes, either true or virtual, our method preserves the model's inherent safety mechanisms while achieving competitive downstream task performance. Extensive experiments confirm its effectiveness and efficiency across multiple benchmarks.

Looking forward, we plan to extend LookAhead Tuning in two directions. First, we aim to explore broader architectures, including multimodal LLMs, and test robustness under diverse adversarial settings. Second, we will investigate automated strategies for optimizing the trade-off between task utility and safety alignment, potentially adapting the preview length or prefix type dynamically.
\section*{Acknowledgments}
This work was supported by the Information Technology Center and State Key Lab of CAD\&CG, Zhejiang University, Ant Group and Zhejiang University - Ant Group Joint Laboratory of Knowledge Graph, Ningbo Natural Science Foundation (2024J020), and the Ministry of Education, Singapore, under the Academic Research Fund Tier 1 (FY2023) (Grant A-8001996-00-00). 

\section*{Ethics Statement}
This work adheres to ethical AI research guidelines, striving to ensure safety in model fine-tuning.
Although our approach has proven effective in reducing harmful output, we recognize the need to continuously evaluate edge cases and unexpected behaviors in real-world deployment.
Moreover, we categorically denounce any malicious misuse of this technology, striving to ensure that the development of AI consistently aligns with human ethical principles.


\bibliographystyle{ACM-Reference-Format}
\bibliography{sample-base}

@String{Computer = "{IEEE} Computer" }

@article{ji2025first,
  title={The First Few Tokens Are All You Need: An Efficient and Effective Unsupervised Prefix Fine-Tuning Method for Reasoning Models},
  author={Ji, Ke and Xu, Jiahao and Liang, Tian and Liu, Qiuzhi and He, Zhiwei and Chen, Xingyu and Liu, Xiaoyuan and Wang, Zhijie and Chen, Junying and Wang, Benyou and others},
  journal={arXiv preprint arXiv:2503.02875},
  year={2025}
}

@article{DBLP:journals/corr/abs-2303-18223,
  author       = {Wayne Xin Zhao and
                  Kun Zhou and
                  Junyi Li and
                  Tianyi Tang and
                  Xiaolei Wang and
                  Yupeng Hou and
                  Yingqian Min and
                  Beichen Zhang and
                  Junjie Zhang and
                  Zican Dong and
                  Yifan Du and
                  Chen Yang and
                  Yushuo Chen and
                  Zhipeng Chen and
                  Jinhao Jiang and
                  Ruiyang Ren and
                  Yifan Li and
                  Xinyu Tang and
                  Zikang Liu and
                  Peiyu Liu and
                  Jian{-}Yun Nie and
                  Ji{-}Rong Wen},
  title        = {A Survey of Large Language Models},
  journal      = {CoRR},
  volume       = {abs/2303.18223},
  year         = {2023},
  url          = {https://doi.org/10.48550/arXiv.2303.18223},
  doi          = {10.48550/ARXIV.2303.18223},
  eprinttype    = {arXiv},
  eprint       = {2303.18223},
  bibsource    = {dblp computer science bibliography, https://dblp.org}
}

@article{lermen2023lora,
  title={Lora fine-tuning efficiently undoes safety training in llama 2-chat 70b},
  author={Lermen, Simon and Rogers-Smith, Charlie and Ladish, Jeffrey},
  journal={arXiv preprint arXiv:2310.20624},
  year={2023}
}

@article{wei2024assessing,
  title={Assessing the Brittleness of Safety Alignment via Pruning and Low-Rank Modifications},
  author={Wei, Boyi and Huang, Kaixuan and Huang, Yangsibo and Xie, Tinghao and Qi, Xiangyu and Xia, Mengzhou and Mittal, Prateek and Wang, Mengdi and Henderson, Peter},
  journal={arXiv preprint arXiv:2402.05162},
  year={2024}
}

@article{yang2023shadow,
  title={Shadow alignment: The ease of subverting safely-aligned language models},
  author={Yang, Xianjun and Wang, Xiao and Zhang, Qi and Petzold, Linda and Wang, William Yang and Zhao, Xun and Lin, Dahua},
  journal={arXiv preprint arXiv:2310.02949},
  year={2023}
}

@article{zhan2023removing,
  title={Removing rlhf protections in gpt-4 via fine-tuning},
  author={Zhan, Qiusi and Fang, Richard and Bindu, Rohan and Gupta, Akul and Hashimoto, Tatsunori and Kang, Daniel},
  journal={arXiv preprint arXiv:2311.05553},
  year={2023}
}

@article{cobbe2021training,
  title={Training verifiers to solve math word problems},
  author={Cobbe, Karl and Kosaraju, Vineet and Bavarian, Mohammad and Chen, Mark and Jun, Heewoo and Kaiser, Lukasz and Plappert, Matthias and Tworek, Jerry and Hilton, Jacob and Nakano, Reiichiro and others},
  journal={arXiv preprint arXiv:2110.14168},
  year={2021}
}

@article{qi2023fine,
  title={Fine-tuning Aligned Language Models Compromises Safety, Even When Users Do Not Intend To!},
  author={Qi, Xiangyu and Zeng, Yi and Xie, Tinghao and Chen, Pin-Yu and Jia, Ruoxi and Mittal, Prateek and Henderson, Peter},
  journal={arXiv preprint arXiv:2310.03693},
  year={2023}
}

@article{huang2024vaccine,
  title={Vaccine: Perturbation-aware Alignment for Large Language Model},
  author={Huang, Tiansheng and Hu, Sihao and Liu, Ling},
  journal={arXiv preprint arXiv:2402.01109},
  year={2024}
}

@article{qi2024safety,
  title={Safety Alignment Should Be Made More Than Just a Few Tokens Deep},
  author={Qi, Xiangyu and Panda, Ashwinee and Lyu, Kaifeng and Ma, Xiao and Roy, Subhrajit and Beirami, Ahmad and Mittal, Prateek and Henderson, Peter},
  journal={arXiv preprint arXiv:2406.05946},
  year={2024}
}

@article{he2024s,
  title={What's in Your" Safe" Data?: Identifying Benign Data that Breaks Safety},
  author={He, Luxi and Xia, Mengzhou and Henderson, Peter},
  journal={arXiv preprint arXiv:2404.01099},
  year={2024}
}

@inproceedings{yi2024vulnerability,
  title={On the vulnerability of safety alignment in open-access llms},
  author={Yi, Jingwei and Ye, Rui and Chen, Qisi and Zhu, Bin and Chen, Siheng and Lian, Defu and Sun, Guangzhong and Xie, Xing and Wu, Fangzhao},
  booktitle={Findings of the Association for Computational Linguistics ACL 2024},
  pages={9236--9260},
  year={2024}
}

@inproceedings{DBLP:conf/acl/YangPFWCZL24,
  author       = {Zhaorui Yang and
                  Tianyu Pang and
                  Haozhe Feng and
                  Han Wang and
                  Wei Chen and
                  Minfeng Zhu and
                  Qian Liu},
  editor       = {Lun{-}Wei Ku and
                  Andre Martins and
                  Vivek Srikumar},
  title        = {Self-Distillation Bridges Distribution Gap in Language Model Fine-Tuning},
  booktitle    = {Proceedings of the 62nd Annual Meeting of the Association for Computational
                  Linguistics (Volume 1: Long Papers), {ACL} 2024, Bangkok, Thailand,
                  August 11-16, 2024},
  pages        = {1028--1043},
  publisher    = {Association for Computational Linguistics},
  year         = {2024},
  url          = {https://doi.org/10.18653/v1/2024.acl-long.58},
  doi          = {10.18653/V1/2024.ACL-LONG.58},
  bibsource    = {dblp computer science bibliography, https://dblp.org}
}

@article{Lin2023TheUS,
  title={The Unlocking Spell on Base LLMs: Rethinking Alignment via In-Context Learning},
  author={Bill Yuchen Lin and Abhilasha Ravichander and Ximing Lu and Nouha Dziri and Melanie Sclar and Khyathi Raghavi Chandu and Chandra Bhagavatula and Yejin Choi},
  journal={ArXiv},
  year={2023},
  volume={abs/2312.01552},
  url={https://api.semanticscholar.org/CorpusID:265608902}
}

@inproceedings{samsum,
    title = "{SAMS}um Corpus: A Human-annotated Dialogue Dataset for Abstractive Summarization",
    author = "Gliwa, Bogdan  and
      Mochol, Iwona  and
      Biesek, Maciej  and
      Wawer, Aleksander",
    booktitle = "Proceedings of the 2nd Workshop on New Frontiers in Summarization",
    month = nov,
    year = "2019",
    address = "Hong Kong, China",
    publisher = "Association for Computational Linguistics",
    url = "https://www.aclweb.org/anthology/D19-5409",
    doi = "10.18653/v1/D19-5409",
    pages = "70--79"
}

@article{Touvron2023Llama2O,
  title={Llama 2: Open Foundation and Fine-Tuned Chat Models},
  author={Hugo Touvron and Louis Martin and Kevin R. Stone and Peter Albert and Amjad Almahairi and Yasmine Babaei and Nikolay Bashlykov and Soumya Batra and Prajjwal Bhargava and Shruti Bhosale and Daniel M. Bikel and Lukas Blecher and Cristian Cant{\'o}n Ferrer and Moya Chen and Guillem Cucurull and David Esiobu and Jude Fernandes and Jeremy Fu and Wenyin Fu and Brian Fuller and Cynthia Gao and Vedanuj Goswami and Naman Goyal and Anthony S. Hartshorn and Saghar Hosseini and Rui Hou and Hakan Inan and Marcin Kardas and Viktor Kerkez and Madian Khabsa and Isabel M. Kloumann and A. V. Korenev and Punit Singh Koura and Marie-Anne Lachaux and Thibaut Lavril and Jenya Lee and Diana Liskovich and Yinghai Lu and Yuning Mao and Xavier Martinet and Todor Mihaylov and Pushkar Mishra and Igor Molybog and Yixin Nie and Andrew Poulton and Jeremy Reizenstein and Rashi Rungta and Kalyan Saladi and Alan Schelten and Ruan Silva and Eric Michael Smith and R. Subramanian and Xia Tan and Binh Tang and Ross Taylor and Adina Williams and Jian Xiang Kuan and Puxin Xu and Zhengxu Yan and Iliyan Zarov and Yuchen Zhang and Angela Fan and Melissa Hall Melanie Kambadur and Sharan Narang and Aur{\'e}lien Rodriguez and Robert Stojnic and Sergey Edunov and Thomas Scialom},
  journal={ArXiv},
  year={2023},
  volume={abs/2307.09288},
  url={https://api.semanticscholar.org/CorpusID:259950998}
}

@article{alber2025medical,
  title={Medical large language models are vulnerable to data-poisoning attacks},
  author={Alber, Daniel Alexander and Yang, Zihao and Alyakin, Anton and Yang, Eunice and Rai, Sumedha and Valliani, Aly A and Zhang, Jeff and Rosenbaum, Gabriel R and Amend-Thomas, Ashley K and Kurland, David B and others},
  journal={Nature Medicine},
  pages={1--9},
  year={2025},
  publisher={Nature Publishing Group US New York}
}

@inproceedings{
zhang2024dissecting,
title={Dissecting learning and forgetting in language model finetuning},
author={Xiao Zhang and Ji Wu},
booktitle={The Twelfth International Conference on Learning Representations},
year={2024},
url={https://openreview.net/forum?id=tmsqb6WpLz}
}

@inproceedings{
lin2024unlocking,
title={The Unlocking Spell on Base {LLM}s:  Rethinking Alignment via In-Context Learning},
author={Bill Yuchen Lin and Abhilasha Ravichander and Ximing Lu and Nouha Dziri and Melanie Sclar and Khyathi Chandu and Chandra Bhagavatula and Yejin Choi},
booktitle={The Twelfth International Conference on Learning Representations},
year={2024},
url={https://openreview.net/forum?id=wxJ0eXwwda}
}

@article{xu2024safedecoding,
  title={SafeDecoding: Defending against Jailbreak Attacks via Safety-Aware Decoding},
  author={Xu, Zhangchen and Jiang, Fengqing and Niu, Luyao and Jia, Jinyuan and Lin, Bill Yuchen and Poovendran, Radha},
  journal={arXiv preprint arXiv:2402.08983},
  year={2024}
}

@inproceedings{DBLP:conf/emnlp/Li0FT24,
  author       = {Hongyu Li and
                  Liang Ding and
                  Meng Fang and
                  Dacheng Tao},
  editor       = {Yaser Al{-}Onaizan and
                  Mohit Bansal and
                  Yun{-}Nung Chen},
  title        = {Revisiting Catastrophic Forgetting in Large Language Model Tuning},
  booktitle    = {Findings of the Association for Computational Linguistics: {EMNLP}
                  2024, Miami, Florida, USA, November 12-16, 2024},
  pages        = {4297--4308},
  publisher    = {Association for Computational Linguistics},
  year         = {2024},
  url          = {https://aclanthology.org/2024.findings-emnlp.249},
  bibsource    = {dblp computer science bibliography, https://dblp.org}
}

@article{DBLP:journals/corr/abs-2308-08747,
  author       = {Yun Luo and
                  Zhen Yang and
                  Fandong Meng and
                  Yafu Li and
                  Jie Zhou and
                  Yue Zhang},
  title        = {An Empirical Study of Catastrophic Forgetting in Large Language Models
                  During Continual Fine-tuning},
  journal      = {CoRR},
  volume       = {abs/2308.08747},
  year         = {2023},
  url          = {https://doi.org/10.48550/arXiv.2308.08747},
  doi          = {10.48550/ARXIV.2308.08747},
  eprinttype    = {arXiv},
  eprint       = {2308.08747},
  bibsource    = {dblp computer science bibliography, https://dblp.org}
}

@article{shallowalign,
  title={Shadow alignment: The ease of subverting safely-aligned language models},
  author={Yang, Xianjun and Wang, Xiao and Zhang, Qi and Petzold, Linda and Wang, William Yang and Zhao, Xun and Lin, Dahua},
  journal={arXiv preprint arXiv:2310.02949},
  year={2023}
}

@article{pelrine2023exploiting,
  title={Exploiting Novel GPT-4 APIs},
  author={Pelrine, Kellin and Taufeeque, Mohammad and Zajac, Michal and McLean, Euan and Gleave, Adam},
  journal={arXiv preprint arXiv:2312.14302},
  year={2023}
}

@inproceedings{DBLP:conf/nips/Wei0SBIXCLZ22,
  author       = {Jason Wei and
                  Xuezhi Wang and
                  Dale Schuurmans and
                  Maarten Bosma and
                  Brian Ichter and
                  Fei Xia and
                  Ed H. Chi and
                  Quoc V. Le and
                  Denny Zhou},
  editor       = {Sanmi Koyejo and
                  S. Mohamed and
                  A. Agarwal and
                  Danielle Belgrave and
                  K. Cho and
                  A. Oh},
  title        = {Chain-of-Thought Prompting Elicits Reasoning in Large Language Models},
  booktitle    = {Advances in Neural Information Processing Systems 35: Annual Conference on Neural Information Processing Systems 2022, NeurIPS 2022, New Orleans, LA, USA, November 28 - December 9, 2022},
  year         = {2022},
  url          = {http://papers.nips.cc/paper\_files/paper/2022/hash/9d5609613524ecf4f15af0f7b31abca4-Abstract-Conference.html},
  bibsource    = {dblp computer science bibliography, https://dblp.org}
}

@article{DBLP:journals/corr/abs-2411-14405,
  author       = {Yu Zhao and
                  Huifeng Yin and
                  Bo Zeng and
                  Hao Wang and
                  Tianqi Shi and
                  Chenyang Lyu and
                  Longyue Wang and
                  Weihua Luo and
                  Kaifu Zhang},
  title        = {Marco-o1: Towards Open Reasoning Models for Open-Ended Solutions},
  journal      = {CoRR},
  volume       = {abs/2411.14405},
  year         = {2024},
  url          = {https://doi.org/10.48550/arXiv.2411.14405},
  doi          = {10.48550/ARXIV.2411.14405},
  eprinttype    = {arXiv},
  eprint       = {2411.14405},
  bibsource    = {dblp computer science bibliography, https://dblp.org}
}

@article{muennighoff2025s1,
  title={s1: Simple test-time scaling},
  author={Muennighoff, Niklas and Yang, Zitong and Shi, Weijia and Li, Xiang Lisa and Fei-Fei, Li and Hajishirzi, Hannaneh and Zettlemoyer, Luke and Liang, Percy and Cand{\`e}s, Emmanuel and Hashimoto, Tatsunori},
  journal={arXiv preprint arXiv:2501.19393},
  year={2025}
}

@inproceedings{DBLP:conf/acl/Zhang0ZX24,
  author       = {Linhai Zhang and
                  Jialong Wu and
                  Deyu Zhou and
                  Guoqiang Xu},
  editor       = {Lun{-}Wei Ku and
                  Andre Martins and
                  Vivek Srikumar},
  title        = {{STAR:} Constraint LoRA with Dynamic Active Learning for Data-Efficient
                  Fine-Tuning of Large Language Models},
  booktitle    = {Findings of the Association for Computational Linguistics, {ACL} 2024,
                  Bangkok, Thailand and virtual meeting, August 11-16, 2024},
  pages        = {3519--3532},
  publisher    = {Association for Computational Linguistics},
  year         = {2024},
  url          = {https://doi.org/10.18653/v1/2024.findings-acl.209},
  doi          = {10.18653/V1/2024.FINDINGS-ACL.209},
  bibsource    = {dblp computer science bibliography, https://dblp.org}
}


\end{document}